\documentclass[11pt,a4paper]{article}
\usepackage[hyperref]{emnlp2018}
\usepackage{times}
\usepackage{latexsym}
\usepackage{booktabs}
\usepackage{graphicx}
\usepackage{amssymb}
\usepackage{subfig}
\usepackage{url}
\usepackage{multirow}

\aclfinalcopy 

\usepackage{etoolbox}
\makeatletter
\patchcmd\@combinedblfloats{\box\@outputbox}{\unvbox\@outputbox}{}{%
   \errmessage{\noexpand\@combinedblfloats could not be patched}%
}%
 \makeatother

\newcommand{\nir}[1]{\textcolor{red}{$_{N}$[#1]}}

\newcommand{\eat}[1]{}
\renewcommand\vec[1]{\overrightarrow{#1}}
\newcommand\cev[1]{\overleftarrow{#1}}

\title{Fake Sentence Detection as a Training Task for Sentence Encoding}

\author{Viresh Ranjan\qquad Heeyoung Kwon \qquad Niranjan Balasubramanian\qquad Minh Hoai\\Department of Computer Science\\
Stony Brook University}

\date{}

\begin{document}

\maketitle
\begin{abstract}
    Sentence encoders are typically trained on language modeling tasks with large unlabeled datasets. While these encoders achieve state-of-the-art results on many sentence-level tasks, they are difficult to train with long training cycles. 
    We introduce fake sentence detection as a new training task for learning sentence encoders. We automatically generate fake sentences by corrupting original sentences from a source collection and train the encoders to produce representations that are effective at detecting fake sentences. This binary classification task turns to be quite efficient for training sentence encoders. We compare a basic BiLSTM encoder trained on this task with a strong sentence encoding models (Skipthought and FastSent) trained on a language modeling task. We find that the BiLSTM trains much faster on fake sentence detection (20 hours instead of weeks) using smaller amounts of data (1M instead of 64M sentences). Further analysis shows the learned representations capture many syntactic and semantic properties expected from good sentence representations.
    

\end{abstract}

\section{Introduction}

Unsupervised sentence encoders are often trained on language modeling based tasks where the encoded sentence representations are used to reconstruct the input sentence~\cite{hill2016learning} or generate neighboring sentences~\cite{kiros2015skip,hill2016learning}. The trained encoders produce sentence representations that achieve the best performance on many sentence-level prediction tasks~\cite{hill2016learning}.

However, training encoders using such language modeling based tasks is difficult. Language model prediction over large vocabularies across large contexts often means having large models (at least in the output layers), requiring large training data and long training times.



Instead, we introduce an unsupervised discriminative training task, {\em fake sentence detection}. The main idea is to generate fake sentences by corrupting an original sentence. We use two methods to generate fake sentences:  {\em word shuffling} where we swap the positions of two words at random and {\em word dropping}, where we drop a word at random from the original sentence.
The resulting fake sentences are mostly similar to the original sentences -- a fake sentence differs from its source in at most two word positions. Given an source corpus of unlabeled English sentences, we build a new collection of sentences by creating multiple fake sentences for every sentence in the source corpus. The training task is then to take any given sentence from this new collection as input and predict whether it is a real or fake sentence. 

This training task formulation has two key advantages:
(i) This binary classification task can be modeled with fewer parameters in the output layer and can be trained more efficiently compared to the language modeling training tasks where the output layer has many parameters depending on the vocabulary size.
(ii) The task forces the encoder to track both syntax and semantics. Swapping words, for instance, can not only break syntax, but can also lead to a sentence that is semantically incoherent or less plausible (e.g., ``John reached Chicago.'' versus ``Chicago reached John''). 

We train a bidirectional long short term memory network (BiLSTM) encoder that produces a representation of the input sentence, which is fed to a three-layer feed-forward network for prediction. We then evaluate this trained encoder {\em without any further tuning} on multiple sentence-level tasks and test for syntactic and semantic properties which demonstrate the benefits of fake sentence training.


In summary, this paper makes the following  contributions:
1) Introduces fake sentence detection as an unsupervised training task for learning sentence encoders that can distinguish between small changes in mostly similar sentences.
2) An empirical evaluation on multiple sentence-level tasks showing representations trained on the fake sentence tasks outperform a strong baseline model trained on language modeling tasks, even when training on small amounts of data (1M vs. 64M sentences) reducing training time from weeks to within 20 hours.

\section{Related Work}
Previous sentence encoding approaches can be broadly classified as supervised~\cite{infersent,cer2018universal,marcheggiani2017encoding,wieting2015towards}, unsupervised~\cite{kiros2015skip,hill2016learning} or semi-supervised approaches~\cite{4582,peters2018deep,dai2015semi,socher2011semi}. 
The supervised approaches train the encoders on tasks such as NLI and use transfer learning to adapt the learned encoders to different downstream tasks. The unsupervised approaches extend the skip-gram~\cite{mikolov2013distributed} to the sentence level, and use the sentence embedding to predict the adjacent sentences. Skipthought~\cite{kiros2015skip} uses a BiLSTM encoder to obtain a fixed length embedding for a sentence, and uses a BiLSTM decoder to predict adjacent sentences. Training Skipthought model is expensive, and one epoch of training on the Toronto BookCorpus~\cite{zhu2015aligning} dataset takes more than two weeks~\cite{hill2016learning} on a single GPU. FastSent~\cite{hill2016learning} uses embeddings of a sentence to predict words from the adjacent sentences. A sentence is represented by simply summing up the word representation of all the words in the sentence. FastSent requires 
less training time than Skipthought, but FastSent has worse performance. Semi-supervised approaches train sentence encoders on large unlabeled datasets, and do a task specific adaptation using labeled data.  

In this work, we propose an unsupervised sentence encoder that takes around 20 hours to train on a single GPU, and outperforms Skipthought and FastSent encoders on multiple downstream tasks. Unlike the previous unsupervised approaches, we use the binary task of real versus fake sentence
classification to train a BiLSTM based sentence encoder.

\section{Training Tasks for Encoders}
We propose a discriminative task for training sentence encoders. The key bottleneck in training sentence encoders is the need for large amounts of labeled data. Prior work use language modeling as a training task leveraging unlabeled text data. Encoders are trained to produce sentence representations which are effective at either generating neighboring sentences (e.g., Skipthought~\cite{kiros2015skip} or at least effective at predict the words in the neighboring sentences~\cite{hill2016learning}. The challenge becomes one of balance between model coverage (i.e. the number of output words it can predict) and model complexity (i.e. the number of parameters needed for prediction).

Rather than address the language modeling challenges, we propose a simpler training task that requires making a single prediction over an input sentence. In particular, we propose to learn a sentence encoder by training a sequential model to solve the binary classification task of detecting whether a given input sentence is fake or real. This real-fake sentence classification task would perhaps be trivial if the fake sentences look very different from the real sentences. We propose two simple methods to generate noisy sentences which look {\em mostly} similar to real sentences. We describe the noisy sentence generation strategies in Section~\ref{sec:Fake Sentence Generation}. Thus, we create a labeled dataset of real and fake sentences, and train a sequential model to distinguish between real and fake sentences, which results in a model whose classification layer has far fewer parameters than previous language model based encoders.
 Our model architecture is described in Section~\ref{sec:Real Vs Fake Sentence Classification}.
 \begin{figure}[!t]
\centering
\includegraphics[scale = 0.27]{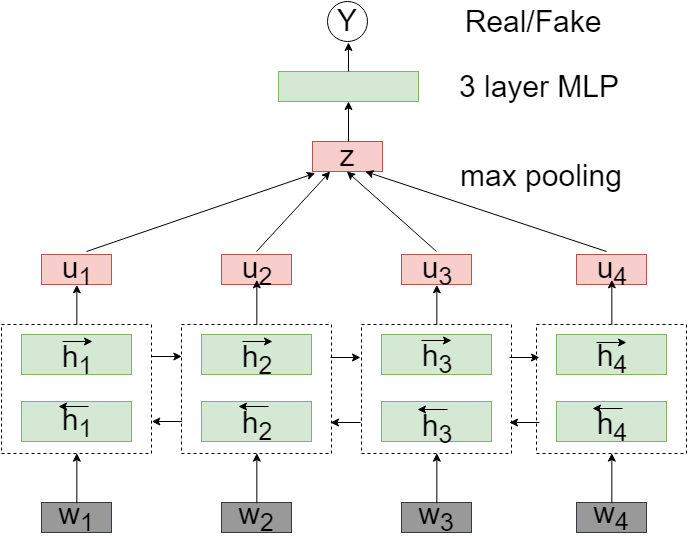}
\vskip -0.1in
\caption{\small Figure shows the block diagram of the encoder and fully connected layers. Encoder consists of a bidirectional LSTM followed by a max pooling layer. For classification, we use a MLP with two hidden layers.\label{encoderBlock}}
\vspace{-1em}
\end{figure}

\subsection{Fake Sentence Generation}\label{sec:Fake Sentence Generation}
For a sentence $X = {w_1, w_2, \ldots, w_n}$ comprising of $n$ words, we consider two strategies to generate a noisy version of the sentence: \textbf{1) WordShuffle}: randomly sample two indices $i$ and $j$ corresponding to words $w_i$ and $w_j$ in $X$, and shuffle the words to obtain the noisy sentence $\hat{X}$. Noisy sentence $\hat{X}$ would be of the same length as the original sentence $X$. \textbf{2) WordDrop}: randomly pick one index $i$ corresponding to word $w_i$ and drop the word from the sentence to obtain $\hat{X}$. Note there can be many variants for this strategy but here we experiment with this basic choice. 

\subsection{Real Versus Fake Sentence Classification}\label{sec:Real Vs Fake Sentence Classification}
Figure~\ref{encoderBlock} shows the proposed architecture of our fake sentence classifier with an encoder and a Multi-layer Perceptron(MLP) with 2 hidden layers. The encoder consists of a bidirectional LSTM followed by a max pooling layer. At each time step we concatenate the forward and backward hidden states to get $u_i = (\vec{h_i}$, $\cev{h_i}$). We apply max-pooling to these concatenated hidden states to get a fixed length representation ($z$), which we then use as input to a MLP for classifying into real/fake classes.

\section{Evaluation Setup}

\textbf{Downstream Tasks:} We compare the sentence encoders trained on a large collection (BookCorpus~\cite{zhu2015aligning}) by testing them on multiple sentence level classification tasks (MR, CR, SUBJ, MPQA, TREC, SST) and one NLI task defined over sentence-pairs (SICK).  We also evaluate the sentence representations for image and caption retrieval tasks on the COCO dataset~\cite{lin2014microsoft}. We use the same evaluation protocol and dataset split as ~\cite{karpathy2015deep,infersent}. Table~\ref{tab:datasets} lists the classification tasks and the datasets. We also compare the sentence representations for how well they capture important syntactic and semantic properties using probing classification tasks~\cite{conneau2018you}. For all downstream and probing tasks, we use the encoders to obtain representation for all the sentences, and train logistic regression classifiers on the training split. We tune the $L_2$-norm regularizer using the validation split, and report the results on the test split.

\begin{table}[!h]
\centering
\small
\begin{tabular}{|c|c|c|c|}
\hline
  Name & Size & Task & Classes  \\ \hline
MR & 11K & Sentiment  & 2  \\ \hline
CR & 4K & Product Review & 2 \\ \hline
TREC & 11K & Question type  & 6  \\ \hline
SST & 70K & Sentiment  & 2  \\ \hline
MPQA & 11K & Opinion Polarity  & 2 \\ \hline
SUBJ & 10K & Subjectivity  & 2 \\ \hline
SICK & 10K & NLI & 3  \\ \hline
COCO &123K & Retrieval & - \\ \hline
\end{tabular}
\vskip -0.1in
\caption{Downstream tasks and datasets.}
\label{tab:datasets}

\end{table}

\textbf{Training Corpus:} The FastSent and Skipthought encoders are trained on the full Toronto BookCorpus of 64M sentences~\cite{zhu2015aligning}. Our models, however, train on a much smaller subset of {\em only} 1M sentences.



\begin{table*}[th!]
\centering
\scalebox{0.83}{
\begin{tabular}{|l|c|c|c|c|c|c|c|c|c|}
\hline
Model             & MR & CR & TREC & SST & MPQA & SUBJ & SICK & COCO-Cap &COCO-Img \\ \hline
FastSent          &  70.8 & 78.4 &  80.6    & -    &  80.6    &   88.7   & -   & - & - \\ \hline
Skipthought (full)&  76.5 & 80.1 & \textbf{92.2}    &  82.0   &   87.1   &   \textbf{93.6}   &   \textbf{82.3} & 72.2 & 66.2  \\
Skipthought (1M)  &  65.2 & 70.9  & 79.2  & 66.9 &  81.6  & 86.1 & 75.6 & - & -
\\ \hline
WordDrop    &  \underline{78.8} & \underline{82.2}   & 86.6  &  \underline{\textbf{82.9}}   &  \underline{\textbf{89.8}}    &  92.7    &  \underline{\textbf{83.2}} & \underline{\textbf{73.8}} & \underline{\textbf{67.3}}    \\ \hline
WordShuffle & \underline{\textbf{79.8}}  & \underline{\textbf{82.4}} &   88.4   &   \underline{\textbf{82.4}}  &  \underline{\textbf{89.8}}    &   92.6   &  \underline{\textbf{82.3}}  & \underline{\textbf{74.2}} & \underline{\textbf{67.3}}  \\ \hline
\end{tabular}}
\caption{Results on downstream tasks: Bold face indicates best result and underlined results show when fake sentence training is better than Skipthought (full). COCO-Cap and COCO-Img are caption and image retrieval tasks on COCO. We report Recall@5 for the COCO
retrieval tasks.}
\vspace{-0.5em}
\label{tab:ClassificationTasks}
\end{table*}

\begin{table*}[h!]
\centering
\scalebox{0.82}{
\begin{tabular}{l|cccccccccc}
\hline
Model             & SentLen & WC & TreeDepth & TopConst & BShift & Tense & SubjNum & ObjNum & SOMO & CoordInv\\ \hline

Skipthought (full)  & 85.4 & 79.6  & 41.1  & \textbf{82.5}  & 69.6  & \textbf{90.4}  & 85.6  & \textbf{83.6}  & 53.9  & 69.1        \\ \hline
Skipthought (1M)      &  54.7 & 33.9 & 30.0  & 60.7 & 58.9 & 85.3 & 76.4 & 70.9 & 51.9 & 61.4  \\ \hline
\hline
WordDrop    & \textbf{86.7} & 90.1  & 48.0  & 81.9 & 73.2  & 87.7  & \textbf{87.3} & 82.7 & 59.2  &  70.6       \\ \hline
WordShuffle & 84.9 & \textbf{91.2} & \textbf{48.8} & 82.3 & \textbf{79.9} & 88.2 & 86.7 & 83.3 & \textbf{59.8} & \textbf{70.7}       \\ \hline

\end{tabular}}
\caption{Probing task accuracies. Tasks: SentLen: predict sentence length, WC: is word in sentence, TreeDepth: depth of syntactic tree, TopConst: predict top-level constituent, BShift: is bigram in flipped in sentence, Tense: predict tense of word, Subj(Obj)Num: singular or plural subject, SOMO: semantic odd man out, CoordInv: is co-ordination is inverted. \label{tab:ProbingTask}
}
\end{table*}

\textbf{Sentence Encoder Implementation:} Our sentence encoder architecture is the same as the BiLSTM-max model~\cite{infersent}. We represent words using 300-d pretrained Glove embeddings~\cite{pennington2014glove}. We use a single layer BiLSTM model, with 2048-d hidden states. The MLP classifier we use for fake sentence detection has two hidden layers with 1024 and 512 neurons.
We train separate models for word drop and word shuffle. The models are trained for 15 epochs with a batch size of 64 using SGD algorithm, when training converges with a validation set accuracy of 87.2 for word shuffle. The entire training completes in less than 20 hours on a single GPU machine.

\textbf{Baseline Approaches:} We compare our results with previous
unsupervised sentences encoders, Skipthought~\cite{kiros2015skip} and FastSent~\cite{hill2016learning}. We use the FastSent and Skipthought results trained on the full BookCorpus as mentioned in~\cite{infersent}.

\section{Results}
\paragraph{Classification and NLI:} Results are shown in Table~\ref{tab:ClassificationTasks}. Both fake sentence training tasks yield better performance on five out of the seven language tasks when compared to Skipthought (full), i.e., even when it is trained on the full BookCorpus. Word drop and word shuffle performances are mostly comparable. The Skipthought (1M) row shows that training on a sentence-level language modeling task can fare substantially worse when trained on a smaller subset of data. FastSent, while easier to train and has faster training cycles, is better than Skipthought (1M) but is worse than the full Skipthought model. 

\paragraph{Image-Caption Retrieval:} On both caption and image retrieval tasks (last 2 columns of Table~\ref{tab:ClassificationTasks}), fake sentence training with word dropping and word shuffle are better than the published Skipthought results. 


\paragraph{Probing Tasks:} Table~\ref{tab:ProbingTask} compares sentence encoders using the recently proposed probing tasks~\cite{conneau2018you}. The goal of each task is to use the input sentence encoding to predict a particular syntactic or semantic property of the original sentence it encodes (e.g., predict if the sentence contains a specific word). Encodings from fake sentence training score higher in six out of the ten tasks. WordShuffle encodings are significantly better than Skipthought 
in some semantic properties: tracking word content~(WC), bigram shuffles (BShift), semantic odd man out (SOMO). Skipthought and WordShuffle are comparable on syntactic properties: agreement (SubjNum, ObjNum, Tense, and CoordInv). The only exception is TreeDepth, where WordShuffle is substantially better. Table~\ref{tab:Bshift} shows examples of the BShift task and cases where the word shuffle and Skipthought models fail. In general we find that word shuffle works better when shifted bigrams involve prepositions, articles, or conjunctions.


\begin{table}[]
\centering
\begin{tabular}{|l|c|c|}
\hline
Shuffled Sentence & WS & ST \\\hline
It shone \underline{the} \underline{in} light . & \checkmark & $\times$  \\ \hline
I seized \underline{the} \underline{and} sword leapt
 & \multirow{2}{*}{\checkmark} & \multirow{2}{*}{$\times$} \\
to the window . &  & \\ \hline
Once again Amadeus held out & \multirow{2}{*}{\checkmark} & \multirow{2}{*}{\checkmark} \\ \underline{arm} \underline{his} . &   &   \\ \hline
When we get inside , I know that & \multirow{2}{*}{$\times$} & \multirow{2}{*}{\checkmark}\\
I have to leave and \underline{Marceline} \underline{find} . &   &   \\ \hline
\end{tabular}

\caption{\small Word shuffle (WS) and Skipthought (ST) performance on BShift. Underlined positions are swapped.\label{tab:Bshift}}

\end{table}

\section{Conclusions}
This work introduced an unsupervised training task, fake sentence detection, where the sentence encoders are trained to produce representations which are effective at detecting if a given sentence is an original or a fake. This leads to better performance on downstream tasks and is able to represent semantic and syntactic properties, while also reducing the amount of training needed. More generally the results suggest that tasks which test for different syntactic and semantic properties in altered sentences can be useful for learning effective representations.
\bibliographystyle{acl_natbib_nourl}
\bibliography{main}
\end{document}